\documentclass[letterpaper, 10 pt, conference]{ieeeconf}  

\IEEEoverridecommandlockouts                              

\usepackage{graphicx} 
\usepackage{epsfig} 
\usepackage{mathptmx} 
\usepackage{times} 
\usepackage{amsmath} 
\usepackage{amssymb}  
\usepackage[capitalize]{cleveref}
\usepackage{array}
\usepackage{multirow}
\usepackage{tikz}
\usepackage{url}
\usepackage[T1]{fontenc}
\usepackage{comment}

\title{\LARGE \bf
AI-Based Multi-Object Relative State Estimation with\\ Self-Calibration Capabilities
}

\author{Thomas Jantos$^{1}$, Christian Brommer$^{1}$, Eren Allak$^{1}$, Stephan Weiss$^{1}$ and Jan Steinbrener$^{1}$
\thanks{$^{1}$The authors are with the Control of Networked Systems Group, University of Klagenfurt, 9020 Klagenfurt am Wörthersee, Austria
        {\tt\small firstname.lastname@ieee.org}}%
\thanks{This work was supported by the Federal Ministry for Climate Action, Environment, Energy, Mobility, Innovation and Technology (BMK) under the grant agreement 881082 (MUKISANO).}%
\thanks{{\textbf{Pre-print version, accepted Jan/2023, DOI follows ASAP~\copyright IEEE.}}}%
}

\begin{document}

\maketitle
\thispagestyle{empty}
\pagestyle{empty}

\begin{abstract}
    The capability to extract task specific, semantic information from raw sensory data is a crucial requirement for many applications of mobile robotics. Autonomous inspection of critical infrastructure with Unmanned Aerial Vehicles (UAVs), for example, requires precise navigation relative to the structure that is to be inspected. Recently, Artificial Intelligence (AI)-based methods have been shown to excel at extracting semantic information such as 6 degree-of-freedom (6-DoF) poses of objects from images. 
    
    In this paper, we propose a method combining a state-of-the-art AI-based pose estimator for objects in camera images with data from an inertial measurement unit (IMU) for 6-DoF multi-object relative state estimation of a mobile robot. The AI-based pose estimator detects multiple objects of interest in camera images along with their relative poses. These measurements are fused with IMU data in a state-of-the-art sensor fusion framework. We illustrate the feasibility of our proposed method with real world experiments for different trajectories and number of arbitrarily placed objects. We show that the results can be reliably reproduced due to the self-calibrating capabilities of our approach.
\end{abstract}

\section{INTRODUCTION}

Mobile robots, such as unmanned aerial vehicles (UAVs), rely on the information of their on-board sensors to autonomously navigate the world. Semantic information, i.e. the higher-level meaning of sensor data, can improve a robot's ability to navigate in its surroundings and allows for more complicated tasks \cite{crespo2020semantic}. In semantic navigation, the robot moves depending on context or task, in many cases with respect to objects of interest in the scene. Such tasks include infrastructure inspection \cite{jordan2018state} or object tracking \cite{li2020keyfilter}. While the goal for the latter is to keep the moving object in the field of view of the camera, infrastructure inspection requires accurate positioning of the robot with respect to a typically static object of interest. Semantic information extracted from the robot's sensor data, namely the detection of the object of interest and its pose relative to the robot are important elements to achieving this task. For example, monitoring power pole insulators for possible damages requires a UAV to fly around the desired insulator and take high resolution images from specific positions to allow for detection of damage or changes over time.

\begin{figure}
    \includegraphics[width=1.0\columnwidth]{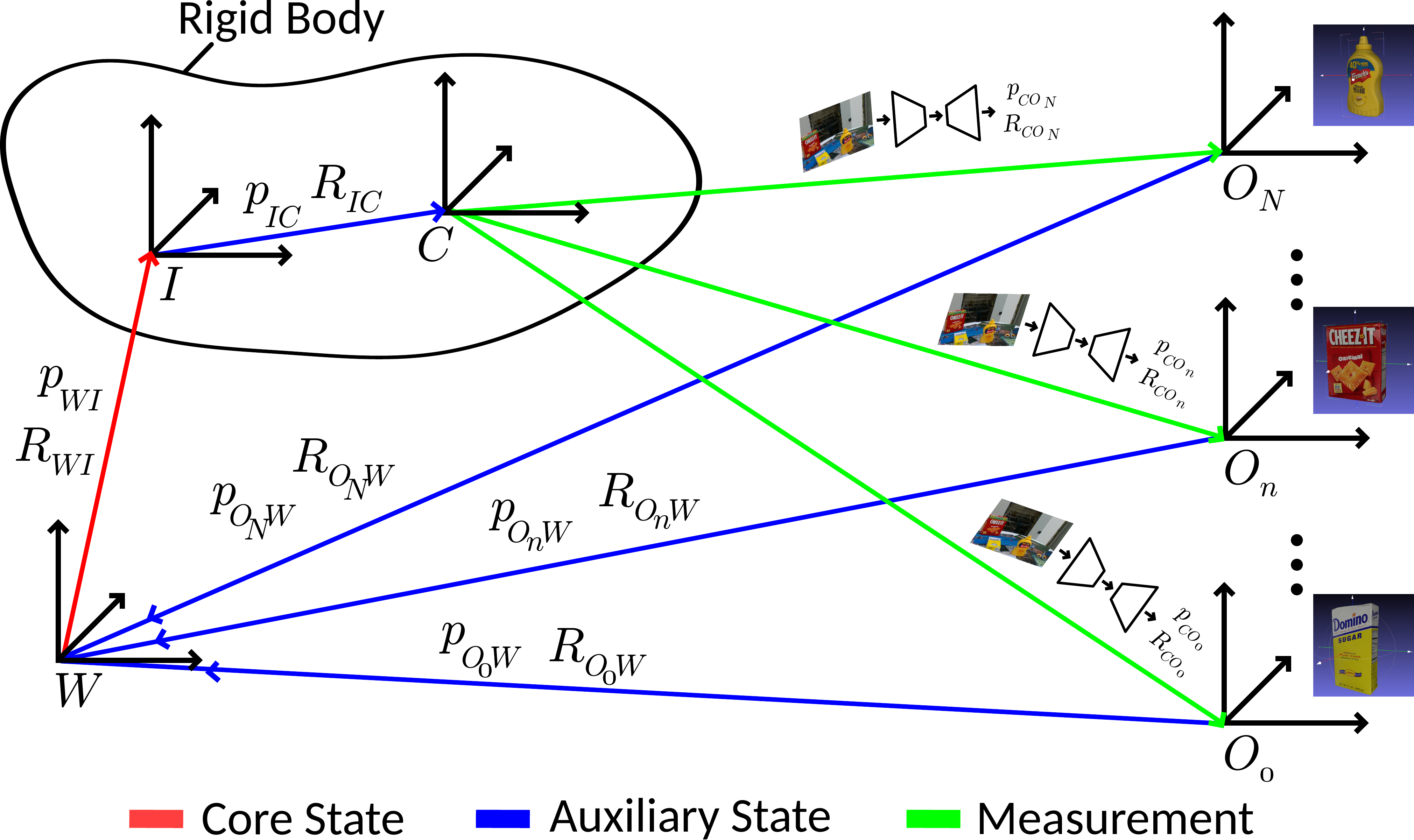}
    \caption{Visualization of the coordinate frames in this work. We estimate the state of a fixed rigid body consisting of IMU $I$ and camera $C$ relative to up to $N$ different objects $O_k$ with respect to a fixed but arbitrary navigation world $W$. In addition to the core states (red), we also estimate the calibration between IMU and camera (blue). We also estimate the pose of the object frames with respect to the world (blue). Our pose sensor consists of AI-based 6-DoF relative pose measurements between camera and objects (green).}
    \label{fig:coordinate_system}
    \vspace{-0.5cm}
\end{figure}

Current autonomous mission execution is typically based on global navigation satellite system (GNSS) for localization of the UAV. GNSS always provides a global position and not a relative position with respect to an object of interest. Moreover, the accuracy is too low for precise, centimeter-range navigation and GNSS is prone to signal loss in proximity to large structures. In this case often other sensor modalities are considered, e.g. visual-inertial odometry (VIO) \cite{weiss2011monocular}. With VIO, a local pose can be estimated by combining the movement of geometrical features (edges, corners) in monocular camera images with data from an inertial measurement unit (IMU). Classical, feature extraction based algorithms are not well suited for semantic navigation as they rely on raw features that do not provide information about any objects in the scene and they struggle with fast or slow motion \cite{Allak2018}.

Recent advancements in artificial intelligence (AI) led to a breakthrough in the extraction of semantic information from raw sensor measurements like path detection using camera images \cite{giusti2015machine}, object recognition with laser scanners \cite{dominguez2011lidar}, semantic segmentation for scene understanding \cite{hofmarcher2019visual}, and recently, 6 degree-of-freedom (6-DoF) pose estimation of objects for robotic grasping \cite{xiang2018posecnn}. Furthermore, the availability of AI capable edge computing devices enables the usage of such methods on mobile robots.

In this paper, we investigate the suitability of AI-based pose estimation for full 6-DoF, object relative state estimation for mobile robotics. We consider a minimal sensor configuration consisting of a single monocular camera and an IMU in line with size, weight and computational power constraints of mobile robotic platforms such as UAVs. We utilize an AI-based pose estimator to detect, classify and estimate the 6-DoF poses of objects of interest contained in each camera image, and then fuse the information with IMU measurements in a state-of-the-art sensor fusion framework to infer the 6-DoF object-relative pose of the robot. A schematic overview of our approach is presented in \cref{fig:coordinate_system}. Our main contributions can be summarized as follows:

\begin{itemize}
    \item Extracting semantic information from images with AI and fusing this relative pose information with IMU data for accurate, 6-DoF, object relative state estimation. 
    \item Formulating a filter-based method to estimate the state of the mobile robot and the pose of multiple, different objects based on 6-DoF relative pose measurements.
    \item Providing a self-calibrating formulation of the filter that does not require any assumptions about the global or relative positions of the different objects in a scene. 
    \item Validating the proposed approach with several real world experiments using objects of a popular 6-DoF object pose challenge data set to show that our method works for different trajectories and different number of objects with reproducible performance.
\end{itemize}

The remainder of the paper is organized as follows. In \cref{sec:related_work}, we summarize the related work. In \cref{sec:method}, we present how we integrate object relative pose measurements into a state estimation framework. In \cref{sec:results}, the experiments and the corresponding results are discussed. Finally, the paper is concluded in \cref{sec:conclusion}. 

\section{RELATED WORK}
\label{sec:related_work}

For state estimation in mobile robotics, typically IMU data and one or more pose sensors such as GNSS are fused together. GNSS provides global position information but not 3D orientation information. In the absence of GNSS signals, VIO, the combination of a monocular camera and IMU data, can estimate the pose of a robot by triangulating the position of the camera given geometrical features from an image and estimating the remaining scale factor with inertial data \cite{weiss2011monocular, bloesch2015robust}. Fusing multiple sensors yields a more robust and reliable estimate of the robot's state. There exist mainly two different approaches for sensor fusion: filter-based, recursive and optimization-based methods. The latter can yield more accurate state estimates but is computationally more demanding due to optimizing across several sensor measurements \cite{sola2022wolf}. In comparison, filtering-based methods, such as Extended Kalman Filters (EKF) \cite{brommer2020mars, moore2016generalized}, are computationally more efficient and thus are well-suited for mobile robotics.

GNSS and classical VIO do not provide object relative pose measurements and thus are not suitable for object relative state estimation. However, image-based 6-DoF relative object pose estimation methods can be utilized as a pose sensor in state estimation frameworks. There exist classical approaches and AI methods based on deep learning. Classical approaches are either template-based, where the object pose is determined by finding a matching template for the current image \cite{cao2016real}, or feature-based, where keypoints are extracted from the image and then matched to the 3D object model \cite{pavlakos20176}. On the other hand, deep learning-based approaches are mostly end-to-end learned methods, where the 6-DoF pose is directly estimated from the input image using convolutional neural networks (CNNs). Deep learning-based methods can be divided even further depending on the amount of additional information used. \cite{xiang2018posecnn, amini2021t6d} take a single RGB image as input to their network and employ symmetry aware losses during training that make use of 3D object model information. 3D object models can also be provided as an additional input to the network \cite{billings2019silhonet}, utilized for refining an initial pose estimate \cite{kehl2017ssd, li2018deepim, li2019cdpn} or for matching keypoints, which were regressed by the network \cite{wang2021gdr}. Other forms of additional information consist of taking multiple images \cite{li2018unified,labbe2020cosypose} or depth maps \cite{krull2015learning, michel2017global}. Recently, we have proposed PoET \cite{jantos2023poet} a 6-DoF multi-object pose estimation framework that achieves state-of-the-art results on benchmark datasets and only takes a single RGB image as input and does not require any additional information during training or inference.

An alternative to object relative state estimation is simultaneous localization and mapping (SLAM) on an object level. \cite{salas2013slam++} uses depth images to extract 6-Dof object pose information. Similar to us, \cite{merrill2022symmetry} use an AI-based pose estimator to predict 6-DoF relative object poses from images. Both approaches fuse the 6-DoF relative pose information from multiple view points together and combine it with graph optimization to estimate the pose of the camera and objects with respect to a map. However, they do not use any other sensors, such as IMU, in their approach. \cite{bowman2017probabilistic} combines IMU measurements, geometric features from images, and 6-DoF object poses in a SLAM approach. Graph optimization still needs to be performed for a graph containing all object poses. In general, the requirement for a map and optimization in SLAM results in a higher computational load for the mobile robot. Our approach still allows for object relative state estimation without this requirement.

Object relative state estimation for mobile robotics has been shown in \cite{thomas2015visual}, where a UAV localizes itself with respect to cylinder shaped infrastructure by extracting geometrical features from images and assuming a known radius. Similarly, \cite{loianno2018localization} used a color-based ellipse-detection algorithm to first detect the object of interest in the image and then used the knowledge about object size, visual appearance and camera parameters to calculate the relative pose to the object. Meanwhile, \cite{mathe2016vision} investigated different classical 6-DoF object pose estimation approaches for object-relative state estimation. They also investigated the use of machine learning to detect the presence of objects by training simple classifiers on classical features extracted from images.

In contrast to that, we propose here a fully AI-based method to extract semantic information from camera images. We do not need to define a geometric model, keypoints or templates to map object appearances in images to relative 6-DoF poses. Moreover, AI-based models are not limited to specific geometric object shapes and remove the need for handcrafted features. In our previous work \cite{jantos2023poet}, we have introduced PoET for 6-DoF pose estimation of objects in RGB images using state-of-the-art AI methods. We mainly focused on the definition, the training, a thorough ablation study and comparison to other deep learning-based methods on benchmark datasets for 6-DoF multi-object pose estimation. In this work, we present a detailed investigation of the suitability of our AI-based object pose estimator as pose sensor for 6-DoF object-relative state estimation of a mobile robot using a state-of-the-art sensor fusion framework with multiple real world experiments.

\section{METHOD}
\label{sec:method}

In this section, we present the design of our approach. First, we explain the notation used for the measurement equations and transformations of coordinate frames. Second, we reason about the choice of frameworks for 6-DoF pose estimation and state estimation. Finally, we describe how the estimated 6-DoF of several known objects can be combined to estimate the 6-DoF pose of the robot. This includes a detailed description of how our choice of sensor fusion algorithm is extended to include 6-DoF pose measurements of each individual object.

\subsection{Notation}
\label{subsec:notation}

Throughout this paper we use the following notation: given three coordinate frames $A$, $B$ and $C$, the transformation ${}_{{\scriptscriptstyle A}}^{}{\mathbf{T}}_{\scriptscriptstyle BC}$ defines frame $C$ with respect to frame $B$ expressed in frame $A$. If the left subscript $A$ is omitted, the transformation is defined in frame $B$. Furthermore, the transformation ${}_{{\scriptscriptstyle A}}^{}{\mathbf{T}}_{\scriptscriptstyle AB}$ can be split up into two parts namely ${}_{{\scriptscriptstyle A}}^{}{\mathbf{p}}_{\scriptscriptstyle AB}$ and ${\mathbf{R}}_{\scriptscriptstyle AB}$, which describe the translation and rotation respectively. Alternatively, the rotation can also be expressed by a quaternion ${\mathbf{q}}_{\scriptscriptstyle AB}$. Each quaternion $\mathbf{q}$ can be represented by $\mathbf{q} = [\mathbf{q_v}~q_w]^T = [q_x ~ q_y ~ q_z ~ q_w]^T$. The quaternion multiplication is represented by $\otimes$. $\mathbf{I}_3$ and $\mathbf{0}_3$ refer to the identity and the null matrix in $\mathbb{R}^{3x3}$, respectively. $[\omega]_\times$ is the skew-symmetric operator as defined in \cite{sola2017quaternion}.

\subsection{Pose and State Estimation Frameworks}
Mobile robots, in particular UAVs, are subject to payload constraints, which impose not only limitations on the size and amount of sensors a robot can carry, but also on the computational power available for data processing. Hence, the necessity arises for efficient and computationally light algorithms. Therefore, we chose our object pose estimation framework PoET \cite{jantos2023poet} as a 6-DoF pose sensor as it only uses RGB images and does not rely on any depth information and 3D object models, removing the need for additional hardware components and reducing the computational load by not having to process 3D models. In a first step, PoET detects all objects it was trained to detect in an image and also predicts their classes. Afterwards, the predicted bounding boxes and multi-scale feature maps are fed to a transformer architecture to predict the relative, up-to scale 6-DoF pose between the camera and each object. The predicted rotation and translation are unique for non-symmetric objects. For objects with one or more symmetry axes, the rotation or translation for some object poses becomes ambiguous with more than one possible solution. The obvious negative effects of this ambiguity on the pose estimation of the robot can be minimized by considering multiple objects in heterogeneous configuration and fusing individual measurements in a proper sensor fusion framework. This mimics an inspection workflow where typically several distinct parts of interest of the structure to be inspected are visible at the same time.

For the sensor fusion framework, we use MaRS \cite{brommer2020mars} for multi-sensor fusion and state estimation due to being lightweight and computationally efficient as it was developed specifically with mobile robotics in mind. MaRS was designed for modularity and separates the propagation of the core state variables based on inertial data from the state updates based on the measurements of the individual sensors. It also uses abstract sensor classes that are type agnostic. This allows for straightforward integration of new sensor modules. For our method, we define a multi-pose sensor, where a single measurement consists of a single RGB image. From each image, we then extract the 6-DoF relative poses of all detected objects with PoET and use them for the EKF update step as described below. 

\subsection{EKF State and Update}

As reference frame for the mobile robotic system, we chose the frame of its IMU ($I$). Thus, our goal is to estimate the pose of the IMU ($I$) with respect to the world ($W$) by measuring the 6-DoF relative poses between the camera ($C$) and a set of objects ($O_k)$. The different coordinate frames are visualized in \cref{fig:coordinate_system}. As mentioned earlier, the relative poses of the objects with respect to the camera are extracted from the image by our AI-based pose estimator dubbed PoET. PoET will only consider objects that it was trained for. Given a single RGB image, a 6-DoF pose ${\mathbf{T}}_{\scriptscriptstyle CO_k}$ is predicted for each detected object of interest and the assignment of the predicted pose to an object is based on the predicted class. For details about the architectural choices in PoET, we refer the reader to \cite{jantos2023poet}. Depending on the total number of objects $N$ in a scene, the full state vector $\mathbf{X}$ is then defined as: 
\begin{align}
 \mathbf{X} = [&{\mathbf{p}}_{\scriptscriptstyle WI}^T, {\mathbf{v}}_{\scriptscriptstyle WI}^T, {\mathbf{q}}_{\scriptscriptstyle WI}^T, {\mathbf{b}}_{\scriptscriptstyle \omega}^T, {\mathbf{b}}_{\scriptscriptstyle a}^T, \\
 &{\mathbf{p}}_{\scriptscriptstyle IC}^T, {\mathbf{q}}_{\scriptscriptstyle IC}^T,{\mathbf{p}}_{\scriptscriptstyle O_0W}^T, {\mathbf{q}}_{\scriptscriptstyle O_0W}^T, \dots, {\mathbf{p}}_{\scriptscriptstyle O_NW}^T, {\mathbf{q}}_{\scriptscriptstyle O_NW}^T]^T \nonumber
\end{align}
The core states necessary for state propagation are the position ${\mathbf{p}}_{\scriptscriptstyle WI}$ of the IMU, its velocity ${\mathbf{v}}_{\scriptscriptstyle WI}$ and its orientation ${\mathbf{q}}_{\scriptscriptstyle WI}$ as well as the gyroscopic bias ${\mathbf{b}}_{\scriptscriptstyle \omega}$ and the accelerometer bias ${\mathbf{b}}_{\scriptscriptstyle a}$. The pose and velocity dynamics are given as \cite{weiss2011monocular}
\begin{align}
    {\Dot{\mathbf{p}}}_{\scriptscriptstyle WI} &= {\mathbf{v}}_{\scriptscriptstyle WI}\\
    {\Dot{\mathbf{v}}}_{\scriptscriptstyle WI} &= {\mathbf{R}}_{\scriptscriptstyle WI} ~ ({\mathbf{a}}_{\scriptscriptstyle m} - {\mathbf{b}}_{\scriptscriptstyle a} - {\mathbf{n}}_{\scriptscriptstyle a}) - \mathbf{g} \\
    {\Dot{\mathbf{q}}}_{\scriptscriptstyle WI} &= \frac{1}{2} \Omega({\mathbf{\omega}}_{\scriptscriptstyle b} - {\mathbf{b}}_{\scriptscriptstyle \omega} - {\mathbf{n}}_{\scriptscriptstyle \omega}) ~ {\mathbf{q}}_{\scriptscriptstyle WI}
\end{align}
where ${\mathbf{a}}_{\scriptscriptstyle m}$ is the measured acceleration in the IMU frame, ${\mathbf{n}}_{\scriptscriptstyle a}$ is the accelerometer noise parameter, $\mathbf{g}$ is the gravity vector in $W$, ${\mathbf{\omega}}_{\scriptscriptstyle b}$ is the measured angular velocity in the IMU frame, ${\mathbf{n}}_{\scriptscriptstyle \omega}$ is the gyroscopic noise parameter, and $\Omega(\omega)$ is the quaternion multiplication matrix of $\omega$. The IMU biases are modeled as random walks.

Furthermore, we estimate the calibration between the IMU and the camera given by ${\mathbf{p}}_{\scriptscriptstyle IC}$ and  ${\mathbf{q}}_{\scriptscriptstyle IC}$.
Due to implementation reasons we assume the number of objects in a scene to be known a priori, but neither the global poses of the objects nor the relative poses between objects is known. Therefore, we additionally estimate an object-world which describes the transformation (${\mathbf{p}}_{\scriptscriptstyle O_kW}, {\mathbf{q}}_{\scriptscriptstyle O_kW}$) between the object frame and the navigation world. Both the camera-IMU extrinsics ${\mathbf{p}}_{\scriptscriptstyle IC}, {\mathbf{q}}_{\scriptscriptstyle IC}$ as well as the object poses ${\mathbf{p}}_{\scriptscriptstyle O_kW}, {\mathbf{q}}_{\scriptscriptstyle O_kW}$ are modeled to remain constant over time.


For each image, every measured relative pose is treated as an independent measurement with which the pose of the camera can be estimated. To calculate the required Jacobians for the update step, we consider the inverted relative pose measurements ${\mathbf{T}}_{\scriptscriptstyle O_kC}$, i.e. the camera pose relative to the object frame. Based on the relationship between the different coordinate frames and the independent relative position $z_{\mathbf{p}_{O_k}}$ and orientation $z_{\mathbf{q}_{O_k}}$ measurements, the residuals for position $\tilde{z}_{\mathbf{p}_{O_k}}$ and orientation $\tilde{z}_{\mathbf{R}_{O_k}}$ can be written as:
%
%
\begin{align}
    \tilde{z}_{\mathbf{p}_{O_k}} &= z_{\mathbf{p}_{O_k}} - \hat{z}_{\mathbf{p}_{O_k}} \nonumber\\ &={\mathbf{p}}_{\scriptscriptstyle O_kC}- ({\mathbf{p}}_{\scriptscriptstyle O_kW} + {\mathbf{R}}_{\scriptscriptstyle O_kW} ({\mathbf{p}}_{\scriptscriptstyle WI} + {\mathbf{R}}_{\scriptscriptstyle WI} ~ {\mathbf{p}}_{\scriptscriptstyle IC}))\\
    \tilde{z}_{\mathbf{R}_{O_k}} &= 2 ~ \frac{\tilde{z}_{\mathbf{q}_{\mathbf{v},O_k}}}{ \tilde{z}_{q_{w,O_k}}} \\
    \tilde{z}_{\mathbf{q}_{O_k}} &= \hat{z}_{\mathbf{q}_{O_k}}^{-1} \otimes z_{\mathbf{q}_{O_k}} = ({\mathbf{q}}_{\scriptscriptstyle O_kW} \otimes {\mathbf{q}}_{\scriptscriptstyle WI} \otimes {\mathbf{q}}_{\scriptscriptstyle IC})^{-1} \otimes {\mathbf{q}}_{\scriptscriptstyle O_kC} \\
    \tilde{z}_{O_k} &= \begin{bmatrix}  \tilde{z}_{\mathbf{p}_{O_k}} \\  \tilde{z}_{\mathbf{R}_{O_k}} \end{bmatrix} \label{eq:res_object}
\end{align}
Given these residuals and depending on a single pose measurement from object $O_k$, the Jacobian for the position $H_\mathbf{p}$ and orientation $H_\mathbf{R}$ with respect to the states is \cite{sola2017quaternion}:
\begin{align}
    H_{\mathbf{p}, {\mathbf{p}}_{\scriptscriptstyle WI}} &= {\mathbf{R}}_{\scriptscriptstyle O_kW} \\
     H_{\mathbf{p},{\mathbf{R}}_{\scriptscriptstyle WI}} &= -{\mathbf{R}}_{\scriptscriptstyle O_kW} {\mathbf{R}}_{\scriptscriptstyle WI} [{\mathbf{p}}_{\scriptscriptstyle IC}]_\times \\
     H_{\mathbf{p}, {\mathbf{p}}_{\scriptscriptstyle IC}} &= {\mathbf{R}}_{\scriptscriptstyle O_kW} {\mathbf{R}}_{\scriptscriptstyle WI} \\
     H_{\mathbf{p}, {\mathbf{p}}_{\scriptscriptstyle O_kW}} &= \mathbf{I}_3\\
     H_{\mathbf{p}, {\mathbf{R}}_{\scriptscriptstyle O_kW}} &= -{\mathbf{R}}_{\scriptscriptstyle O_kW} [{\mathbf{p}}_{\scriptscriptstyle WI}]_\times -{\mathbf{R}}_{\scriptscriptstyle O_kW} [{\mathbf{R}}_{\scriptscriptstyle WI} ~ {\mathbf{p}}_{\scriptscriptstyle IC}]_\times\\
     H_{\mathbf{R},{\mathbf{R}}_{\scriptscriptstyle WI}} &= {\mathbf{R}}_{\scriptscriptstyle IC}^T \\
     H_{\mathbf{R}, {\mathbf{R}}_{\scriptscriptstyle IC}} &= \mathbf{I}_3\\
     H_{\mathbf{R}, {\mathbf{R}}_{\scriptscriptstyle O_kW}} &= {\mathbf{R}}_{\scriptscriptstyle IC}^T {\mathbf{R}}_{\scriptscriptstyle WI}^T
\end{align}
where, e.g. $ H_{\mathbf{p}, {\mathbf{p}}_{\scriptscriptstyle WI}}$ only considers the part of the residual $\tilde{z}_{\mathbf{p}_{O_k}}$ that depends on the state ${\mathbf{p}}_{\scriptscriptstyle WI}$.
The rest of the Jacobians are equal to $\textbf{0}_3$. As relative pose measurements for different objects are independent of each other, the Jacobians for the other ($i \neq n$) object-world states, i.e. $H_{\mathbf{p},{\mathbf{p}}_{\scriptscriptstyle O_iW}}$, $H_{\mathbf{p}, {\mathbf{R}}_{\scriptscriptstyle O_iW}}$, $H_{\mathbf{R}, {\mathbf{p}}_{\scriptscriptstyle O_iW}}$, $H_{\mathbf{R}, {\mathbf{R}}_{\scriptscriptstyle O_iW}}$, are all equal to $\mathbf{0}_3$. For a single object $O_k$, the Jacobian is given by stacking the individual components:
\begin{align}
    H_{\mathbf{p}, O_k} = [&H_{\mathbf{p}, {\mathbf{p}}_{\scriptscriptstyle WI}} \; H_{\mathbf{p},{\mathbf{v}}_{\scriptscriptstyle WI}} \; H_{\mathbf{p},{\mathbf{R}}_{\scriptscriptstyle WI}} \; H_{\mathbf{p},{\mathbf{b}}_{\scriptscriptstyle \omega}} \; H_{\mathbf{p},{\mathbf{b}}_{\scriptscriptstyle a}} \\
    & H_{\mathbf{p}, {\mathbf{p}}_{\scriptscriptstyle IC}} \; H_{\mathbf{p}, {\mathbf{R}}_{\scriptscriptstyle IC}} \; H_{\mathbf{p}, {\mathbf{p}}_{\scriptscriptstyle O_0W}} \;  H_{\mathbf{p}, {\mathbf{R}}_{\scriptscriptstyle O_0W}} \nonumber \\
    & \dots H_{\mathbf{p}, {\mathbf{p}}_{\scriptscriptstyle O_NW}} \;  H_{\mathbf{p}, {\mathbf{R}}_{\scriptscriptstyle O_NW}}] \nonumber
\end{align}
\begin{align}
    H_{\mathbf{R}, O_k} = [&H_{\mathbf{R}, {\mathbf{R}}_{\scriptscriptstyle WI}} \; H_{\mathbf{R},{\mathbf{v}}_{\scriptscriptstyle WI}} \; H_{\mathbf{R},{\mathbf{R}}_{\scriptscriptstyle WI}} \; H_{\mathbf{R},{\mathbf{b}}_{\scriptscriptstyle \omega}} \; H_{\mathbf{R},{\mathbf{b}}_{\scriptscriptstyle a}} \\
    & H_{\mathbf{R}, {\mathbf{p}}_{\scriptscriptstyle IC}} \; H_{\mathbf{R}, {\mathbf{R}}_{\scriptscriptstyle IC}} \; H_{\mathbf{R}, {\mathbf{p}}_{\scriptscriptstyle O_0W}} \;  H_{\mathbf{R}, {\mathbf{R}}_{\scriptscriptstyle O_0W}} \nonumber \\
    & \dots H_{\mathbf{R}, {\mathbf{p}}_{\scriptscriptstyle O_NW}} \;  H_{\mathbf{R}, {\mathbf{R}}_{\scriptscriptstyle O_NW}}] \nonumber
\end{align}
\begin{equation}
\label{eq:jacobian_object}
    H_{O_k} = \begin{bmatrix} H_{\mathbf{p}, O_k} \\ H_{\mathbf{R}, O_k} \end{bmatrix}
\end{equation}

Depending on the current image, the final residual $z$ and observation matrix $\mathbf{H}$ for the state update is determined by vertically stacking the residuals and Jacobians, individually, from \cref{eq:res_object} and \cref{eq:jacobian_object}, respectively, for each object that was detected for the current update step. Similar to hardware sensors, our AI-based pose sensor might return faulty or inaccurate measurements. In a similar fashion as described in \cite{brommer2020improved}, we conduct a $\chi^2$ test based on the EKF innovation $\mathbf{S}$ and the residual to detect outlier measurements. This is applied for the measurement of each object individually. Outlier measurements for object $O_k$ are then rejected and the final residual and Jacobian have to be rebuild by masking the corresponding rows. The correction is then calculated based on this final total residual and associated Jacobian. In the update step, measurement uncertainties for each measurement can be considered. For the present work, these uncertainties have been fixed to 10 cm and 20 degrees for the position and orientation measurement of each object, respectively. These numbers were determined based on the standard deviation of PoET across a video sequence reported in \cite{jantos2023poet}.

The proper initialization of the individual frames is an important aspect to consider. At the beginning of the recording, we initialize an arbitrary but fixed navigation world $W$. Without loss of generality, the IMU frame is initialized at the origin of $W$. The initial extrinsic calibration between the IMU and camera is determined through visual-inertial calibration \cite{furgale2013unified}. Each object-world is initialized when the corresponding object is seen by the camera for the first time. The object frame is then placed with respect to the world frame by taking the currently estimated pose of the camera and the relative pose measurement:
\begin{align}
    {\mathbf{R}}_{\scriptscriptstyle O_kW} &= {\mathbf{R}}_{\scriptscriptstyle O_kC} {\mathbf{R}}_{\scriptscriptstyle IC}^T {\mathbf{R}}_{\scriptscriptstyle WI}^T\\
    {\mathbf{p}}_{\scriptscriptstyle O_kW} &= {\mathbf{p}}_{\scriptscriptstyle O_kC} - {\mathbf{R}}_{\scriptscriptstyle O_kW} ({\mathbf{R}}_{\scriptscriptstyle WI} ~ {\mathbf{p}}_{\scriptscriptstyle IC} + {\mathbf{p}}_{\scriptscriptstyle WI})
\end{align}
In our problem formulation, the robot's pose $I$ is estimated relative to a set of object-worlds $O_k$ through relative pose measurements. As the robot's pose is relative to a world frame and the measurements are relative to the corresponding object frames, the object-worlds can be placed freely with respect to the world frame, which does cause observability issues. By fixing the state of one object-world reference frame, the system is rendered observable. This fixed object, from now on called the main object $O_m$, serves as the anchor point for the object-relative 6-DoF state estimation. The position of the main object's frame with respect to the navigation world is not changed, i.e. the corresponding Jacobians $H_{\mathbf{p}, {\mathbf{p}}_{\scriptscriptstyle O_mW}}$ and $H_{\mathbf{R}, {\mathbf{p}}_{\scriptscriptstyle O_mW}}$ are set to $\mathbf{0}_3$. Measurements in which the main object is not visible in the picture are directly rejected. Otherwise, estimating the object-world of additional objects while the anchor is not visible leads to ambiguous updates.

The propagation step is performed at the rate of the IMU sensor readings, while the update step happens with the frequency of the camera images.

\section{EXPERIMENTS \& RESULTS}
\label{sec:results}

\begin{figure}
    \centering
    \vspace{0.1cm}
    \includegraphics[width=1.0\columnwidth]{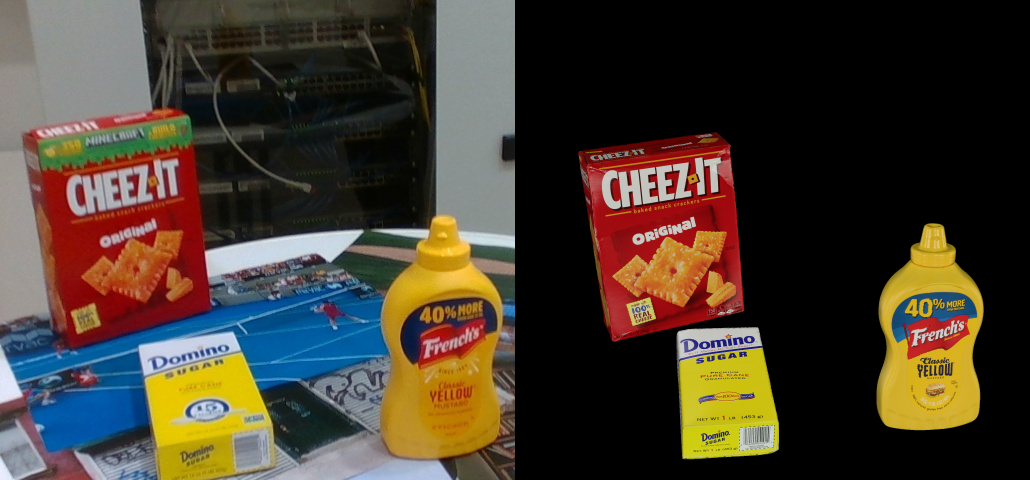}
    \caption{Object configuration that was used for sequence 4 (left) and object poses as estimated by PoET (right). Note the difference in object package coloring between real-world objects (left) and YCB-V objects used for training PoET (right). The left image, as shown here, is directly fed into our pose estimation framework to get the 6-DoF relative pose measurements.}
    \label{fig:example_config}
    \vspace{-0.6cm}
\end{figure}

In this section, we present the experiments conducted and discuss the results in detail. We trained PoET on the YCB-V dataset \cite{xiang2018posecnn} as described in \cite{jantos2023poet}, a benchmark dataset for 6-DoF pose estimation, and chose a subset of objects to serve as objects of interest in our experiments. The images and IMU data were recorded using an Intel Realsense D435i with an RGB resolution of 1280x720 and 30 FPS. After undistorting the images, they were cropped to a resolution of 640x480, which is the standard resolution of the YCB-V dataset. We record our own real data by placing the objects in our motion capture room and moving around the objects with the camera while tracking the body of the camera. This mimics the inspection of a set of objects of interest with a mobile robotic platform with 6-DoFs. An example object configuration and image can be found in \cref{fig:example_config}. Because we do not record any information regarding the global position of the objects in the room, the trajectory derived from the object-relative state estimation has to be aligned with the ground truth trajectory to calculate the error metrics. It is important to note the differences between the benchmark data and our own real data. The camera which was used to record the YCB-V dataset has a different recording resolution, field of view, and focal point than the camera used during our experiments. In addition, some real world objects had slightly different appearance (size and coloring) than the ones used for the data set that PoET was trained with. The results reported here have been obtained with the YCB-V trained model of PoET. We did not perform any retraining or fine-tuning of the model to adapt to these differences.

\begin{table}
    \centering{}\caption{RMSE for position and orientation across the whole trajectory for different sequences.}
\scalebox{0.9}{
\begin{tabular}{c|c|c|c}
Sequence & \#objects & {(}x, y, z{)} {[}m{]} & {(}roll, pitch, yaw{)} {[}deg{]}\tabularnewline
\hline 
\hline 
1 & 3 & {[}0.066, 0.163, 0.034{]} & {[}3.00, 6.81, 4.52{]}\tabularnewline
2 & 3 & {[}0.162, 0.358, 0.233{]}  & {[}54.38, 19.39, 20.27{]}\tabularnewline
3 & 2 & {[}0.163, 0.338, 0.136{]} & {[}25.34, 16.57, 15.13{]}\tabularnewline
4 & 4 & {[}0.045, 0.157, 0.108{]} & {[}15.38, 5.58, 10.02{]}\tabularnewline
5 & 1 & {[}0.075, 0.031, 0.054{]}  & {[}61.12, 9.66, 26.19{]}\tabularnewline
6 & 3 & {[}0.061, 0.117, 0.029{]}  & {[}11.00, 5.77, 7.41{]}\tabularnewline
7 & 4 & {[}0.093, 0.159, 0.099{]}  & {[}19.38, 12.38, 27.24{]}\tabularnewline
8 & 3 & {[}0.127, 0.165, 0.104{]} & {[}26.19, 11.95, 10.70{]}\tabularnewline
\hline 
mean  & \multirow{2}{*}{-} & {[}0.099, 0.186, 0.099{]}  & {[}26.97, 11.0, 15.19{]} \tabularnewline
$\pm$ std &  & $\pm$ {[}0.043, 0.102, 0.062{]} & $\pm$ {[}19.18, 4.75, 8.01{]}\tabularnewline
\end{tabular}
}

    \label{tab:sequences_overview}
    \vspace{-0.3cm}
    \centering{}\caption{RMSE for position and orientation across the whole trajectory for different runs of sequence 4.}
\begin{tabular}{c|c|c}
Run & {(}x, y, z{)} {[}m{]} & {(}roll, pitch, yaw{)} {[}deg{]}\tabularnewline
\hline 
\hline 
1 & {[}0.094, 0.107, 0.119{]} & {[}4.72, 7.08, 13.16{]}\tabularnewline
2 & {[}0.077, 0.121, 0.103{]} & {[}1.33, 7.24, 8.05{]}\tabularnewline
3 & {[}0.049, 0.096, 0.092{]} & {[}3.14, 1.72, 14.49{]}\tabularnewline
4 & {[}0.065, 0.119, 0.101{]} & {[}7.75, 5.25, 12.90{]}\tabularnewline
5 & {[}0.069, 0.119, 0.099{]} & {[}2.79, 3.35, 11.57{]}\tabularnewline
6 & {[}0.047, 0.099, 0.099{]} & {[}2.53, 3.63, 12.21{]}\tabularnewline
7 & {[}0.059, 0.090, 0.088{]} & {[}1.92, 5.01, 10.89{]}\tabularnewline
8 & {[}0.057, 0.093, 0.083{]} & {[}3.68, 2.92, 11.37{]}\tabularnewline
9 & {[}0.074, 0.090, 0.096{]} & {[}6.85, 4.69, 13.11{]}\tabularnewline
10 & {[}0.076, 0.101, 0.118{]} & {[}7.19, 6.02, 12.10{]}\tabularnewline
\hline 
mean & {[}0.067, 0.104, 0.010{]} & {[}4.19, 4.69, 11.99{]}\tabularnewline
$\pm$ std & $\pm$ {[}0.014, 0.012, 0.011{]} & $\pm${[}2.20, 1.71, 1.64{]}
\vspace{-0.3cm}
\end{tabular}
    \label{tab:seq3}
    \vspace{-0.3cm}
    \centering{}\caption{RMSE for position and orientation across the whole trajectory for different runs of sequence 6.}
\begin{tabular}{c|c|c}
Run & {(}x, y, z{)} {[}m{]} & {(}roll, pitch, yaw{)} {[}deg{]}\tabularnewline
\hline 
\hline 
1 & {[}0.061, 0.117, 0.029{]} & {[}11.00, 5.77, 7.41{]}\tabularnewline
2 & {[}0.092, 0.122, 0.052{]} & {[}17.17, 16.97, 7.16{]}\tabularnewline
3 & {[}0.072, 0.112, 0.025{]} & {[}13.08, 24.64, 5.93{]}\tabularnewline
4 & {[}0.083, 0.110, 0.037{]} & {[}12.11, 8.24, 4.58{]}\tabularnewline
5 & {[}0.074, 0.089, 0.023{]} & {[}16.02, 5.99, 6.15{]}\tabularnewline
6 & {[}0.067, 0.079, 0.031{]} & {[}12.29, 11.97, 7.19{]}\tabularnewline
7 & {[}0.064, 0.126, 0.025{]} & {[}16.64, 12.26, 7.72{]}\tabularnewline
8 & {[}0.059, 0.093, 0.036{]} & {[}15.86, 6.31, 6.41{]}\tabularnewline
9 & {[}0.067, 0.092, 0.046{]} & {[}12.38, 4.69, 4.49{]}\tabularnewline
10 & {[}0.057, 0.150, 0.031{]} & {[}12.95, 16.12, 6.89{]}\tabularnewline
\hline 
mean & {[}0.069, 0.109, 0.034{]} & {[}13.95, 11.29, 6.39{]}\tabularnewline
$\pm$ std & $\pm$ {[}0.010, 0.020, 0.009{]} & $\pm${[}2.11, 6.09, 1.07{]}\tabularnewline
\end{tabular}
    \label{tab:seq5}
    \vspace{-0.7cm}
\end{table}

\begin{figure*}
    \centering
    \includegraphics[width=1.0\columnwidth]{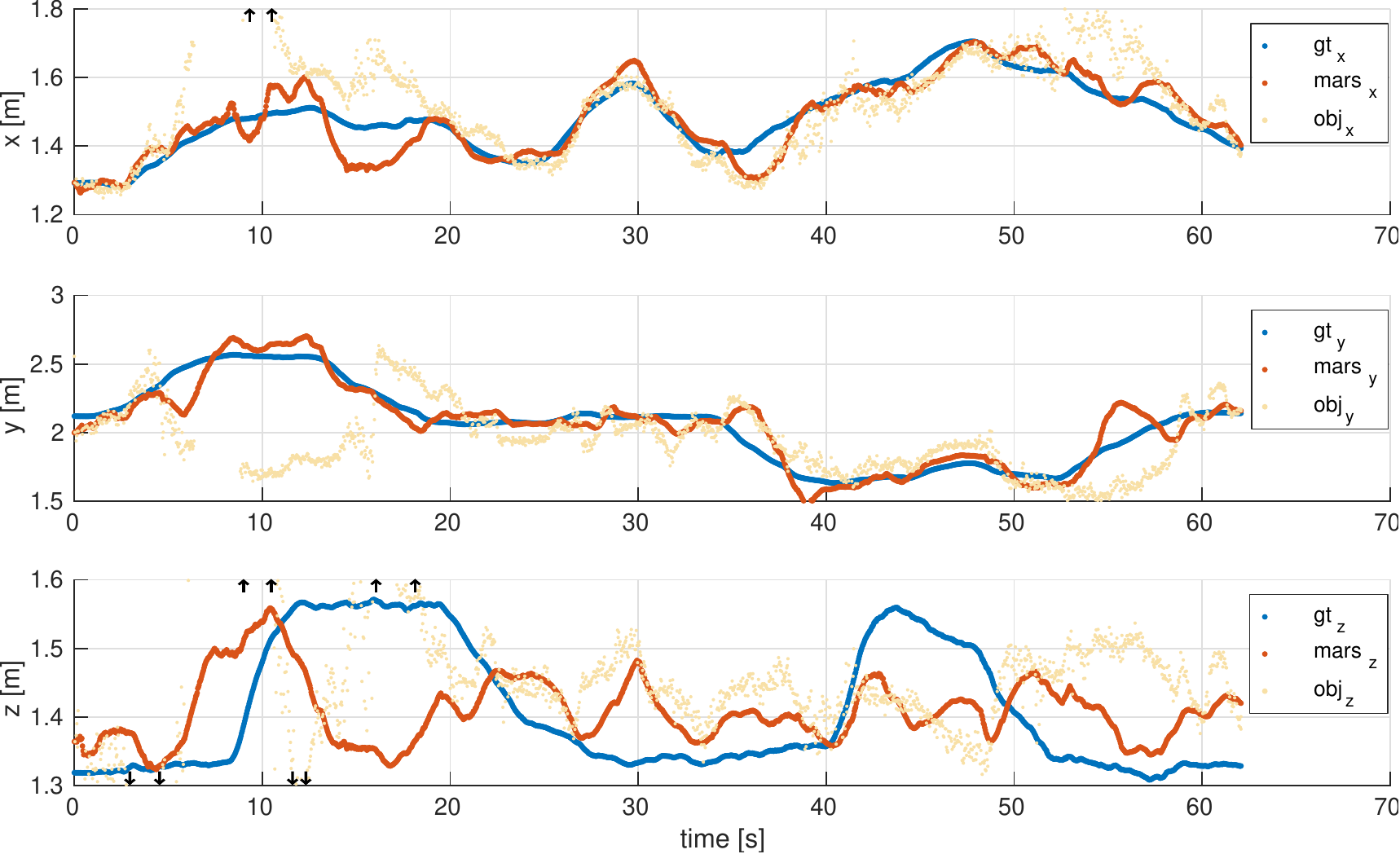}
    \includegraphics[width=1.0\columnwidth]{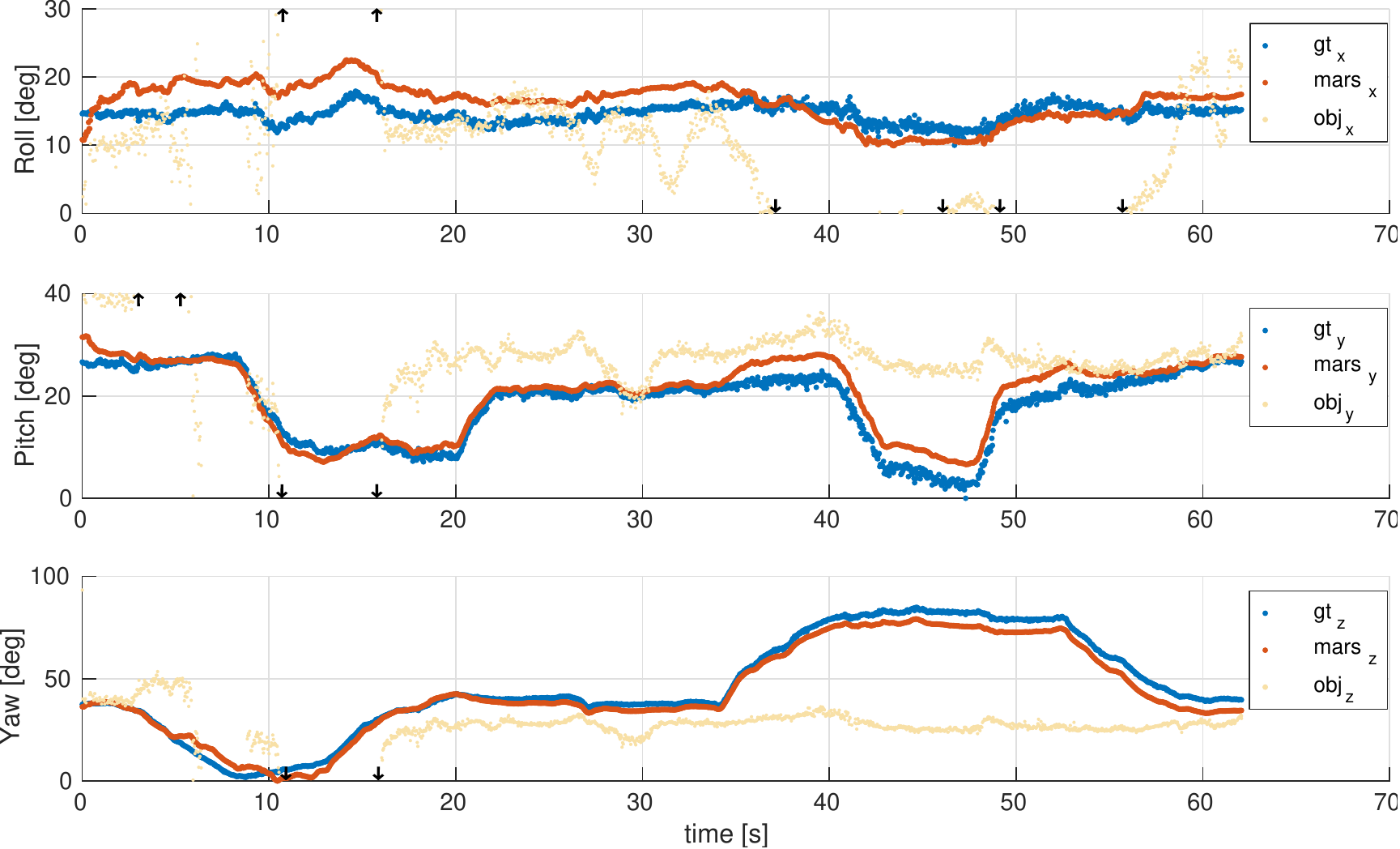}
    \caption{Comparison of estimated position and orientation in Euler angles (mars) and the ground (gt) for run 8 of sequence 4. The components of the position (x, y, z) and orientation (roll, pitch, yaw) are plotted individually for the whole sequence. Additionally, we compare the reprojected IMU pose given the raw PoET estimates for object 3 (obj). The black arrows enclose a section in which the reprojected IMU pose is out of plotting range. Important to note, the object was not visible in the camera images between 6.4s and 8.8s.}
    \label{fig:example_trajectory}
    \vspace{-0.6cm}
\end{figure*}

We conducted two different experiments. First, we investigated the performance of our approach for 8 different sequences. The sequences varied with respect to the number of objects present, the constellation of the objects and the trajectory performed by the camera around the objects. We calculate the root mean square error (RMSE) for the position and Euler angles by comparing the estimated trajectory with the respective ground truth trajectory for a single recording of that sequence. Moreover, we calculate the average RMSE and the standard deviation (std) over all trajectories. The results are summarized in \cref{tab:sequences_overview}. The overall performance across all sequences shows that our method is able to sufficiently estimate the state given AI-based, relative 6-DoF pose measurements for a variety of scenarios. For most sequences, the position RMSE is around 10 cm or less, which is an acceptable error for this rather complex task. Furthermore, these results indicate that our method is applicable for object relative state estimation. In some cases, i.e. sequence 2 and 3, the achieved performance is worse than in others. This is due to outliers in the predictions of the 6-DoF pose estimator. Objects being partially out of the image or ambiguous viewpoints as well as motion blur in the images lead to wrong pose estimates. Especially the latter one results in the predictions for all objects in a single image to be wrong. While the $\chi^2$ test helps to reject such frames, multiple consecutive images with noisy or wrong measurements will still affect the state estimation as it can result in phases with no updates and only IMU propagation leading to deadreckoning. In such cases, a prediction of uncertainties for each object and measurement would lead to better results rather than working with the fixed values described above. Integration of aleatoric and epistemic uncertainties for the predictions of PoET is subject to future work.

To illustrate the reproducibility of our approach, we chose two out of the 8 sequences (sequence 4 and sequence 6) and repeated each sequence 10 times resulting in similar but not exactly the same trajectories. For each sequence, the RMSE across the whole trajectory for each run and the average RMSE and std across all runs are summarized in \cref{tab:seq3} and \cref{tab:seq5}, respectively. For both sequences, we are able to reproduce the performance of our method across 10 independent runs with a low standard deviation. This shows an AI-based component can be reliably incorporated into the state estimation of a robot.

\begin{figure}
    \centering
    \includegraphics[width=1.0\columnwidth]{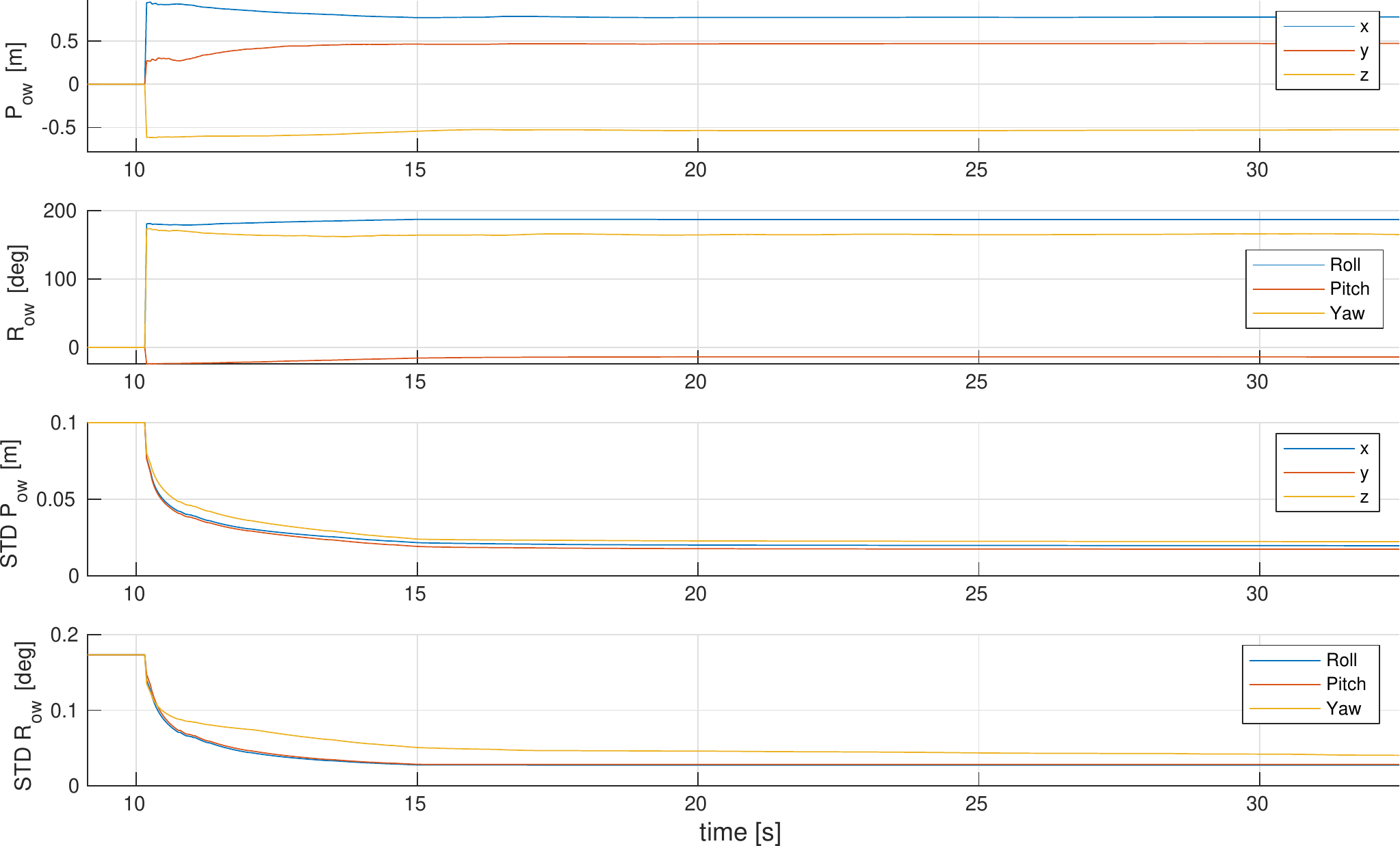}
    \caption{Visualization of the estimated object-world state (${\mathbf{p}}_{\scriptscriptstyle O_kW}, {\mathbf{q}}_{\scriptscriptstyle O_kW}$) and the corresponding state covariance represented by the std for a non-main object for run 8 of sequence 4. The position is split up into (x, y, z), while the orientation is represented by the Euler angles. The states are plotted from the point of time the object is first observed (at about 10s) until the states converge. At the beginning the object state is wrongly initialized due to perhaps a noisy measurement. However, after about 5 seconds the state converges and the uncertainty becomes minimal. }
    \label{fig:obj_states}
    \vspace{-0.7cm}
\end{figure}

Moreover, we compare the estimated and the ground truth position and orientation across the whole trajectory for an example recording in \cref{fig:example_trajectory}. This example shows that our approach reliably estimates the position and orientation for the whole duration of the recording. Furthermore, the graphs show that the raw measurements of a single object sometimes lead to a reprojected IMU pose that does not align with the ground truth trajectory. However, by fusing IMU information with pose measurements from multiple objects our method is able to reliably estimate the trajectory, despite outlier measurements. In addition to that, we show in \cref{fig:obj_states} an example for the self-calibration capabilities of our approach with respect to the object-world states. The object-world is wrongly initialized after it was first observed due to a possible noisy measurement. Nonetheless, the state converges after 5 seconds. 

\section{CONCLUSIONS}
\label{sec:conclusion}

In this paper, we investigated object relative state estimation for mobile robots with an AI-based method to extract semantic information (object class and pose) from single RGB images. We defined a minimal sensor configuration, consisting of an RGB camera and IMU, and an experimental scenario in which object relative state estimation is required, mimicking the task of inspection and monitoring. We derived and implemented a filter-based solution for full state estimation of a mobile robot given 6-DoF relative pose measurements. Additionally, our method does not require any initial information about the global and relative poses of the objects. By defining object-world states, the coordinate frame of each object is estimated concurrently with respect to a common navigation world by using one of the objects as an anchor point. Our experiments with own real data showed that our method can be used for state estimation of the mobile robot in different scenarios and that the results can be reliably reproduced. Our results show that AI-based, semantic information from a single sensor is sufficient in combination with IMU data for accurate state estimation. Future work will consider incorporating aleatoric and epistemic uncertainties of the AI-based predictions in the sensor fusion framework for improved outlier rejection as well the integration of our proposed approach on a real UAV for closed loop experiments.

\bibliographystyle{IEEEtran.bst}
\bibliography{root.bib}

\begin{thebibliography}{10}
\providecommand{\url}[1]{#1}
\csname url@rmstyle\endcsname
\providecommand{\newblock}{\relax}
\providecommand{\bibinfo}[2]{#2}
\providecommand\BIBentrySTDinterwordspacing{\spaceskip=0pt\relax}
\providecommand\BIBentryALTinterwordstretchfactor{4}
\providecommand\BIBentryALTinterwordspacing{\spaceskip=\fontdimen2\font plus
\BIBentryALTinterwordstretchfactor\fontdimen3\font minus
  \fontdimen4\font\relax}
\providecommand\BIBforeignlanguage[2]{{%
\expandafter\ifx\csname l@#1\endcsname\relax
\typeout{** WARNING: IEEEtran.bst: No hyphenation pattern has been}%
\typeout{** loaded for the language `#1'. Using the pattern for}%
\typeout{** the default language instead.}%
\else
\language=\csname l@#1\endcsname
\fi
#2}}

\bibitem{crespo2020semantic}
J.~Crespo, J.~C. Castillo, O.~M. Mozos, and R.~Barber, ``Semantic information
  for robot navigation: A survey,'' \emph{Applied Sciences}, vol.~10, no.~2, p.
  497, 2020.

\bibitem{jordan2018state}
S.~Jordan, J.~Moore, S.~Hovet, J.~Box, J.~Perry, K.~Kirsche, D.~Lewis, and
  Z.~T.~H. Tse, ``State-of-the-art technologies for uav inspections,''
  \emph{IET Radar, Sonar \& Navigation}, vol.~12, no.~2, pp. 151--164, 2018.

\bibitem{li2020keyfilter}
Y.~Li, C.~Fu, Z.~Huang, Y.~Zhang, and J.~Pan, ``Keyfilter-aware real-time uav
  object tracking,'' in \emph{2020 IEEE International Conference on Robotics
  and Automation (ICRA)}.\hskip 1em plus 0.5em minus 0.4em\relax IEEE, 2020,
  pp. 193--199.

\bibitem{weiss2011monocular}
S.~Weiss, D.~Scaramuzza, and R.~Siegwart, ``Monocular-slam--based navigation
  for autonomous micro helicopters in gps-denied environments,'' \emph{Journal
  of Field Robotics}, vol.~28, no.~6, pp. 854--874, 2011.

\bibitem{Allak2018}
E.~Allak, A.~Hardt-Stremayr, and S.~Weiss, ``Key-frame strategy during fast
  image-scale changes and zero motion in {VIO} without persistent features,''
  in \emph{2018 {IEEE}/{RSJ} International Conference on Intelligent Robots and
  Systems ({IROS})}.\hskip 1em plus 0.5em minus 0.4em\relax {IEEE}, oct 2018.

\bibitem{giusti2015machine}
A.~Giusti, J.~Guzzi, D.~C. Cire{\c{s}}an, F.-L. He, J.~P. Rodr{\'\i}guez,
  F.~Fontana, M.~Faessler, C.~Forster, J.~Schmidhuber, G.~Di~Caro,
  \emph{et~al.}, ``A machine learning approach to visual perception of forest
  trails for mobile robots,'' \emph{IEEE Robotics and Automation Letters},
  vol.~1, no.~2, pp. 661--667, 2015.

\bibitem{dominguez2011lidar}
R.~Dominguez, E.~Onieva, J.~Alonso, J.~Villagra, and C.~Gonzalez, ``Lidar based
  perception solution for autonomous vehicles,'' in \emph{2011 11th
  International Conference on Intelligent Systems Design and
  Applications}.\hskip 1em plus 0.5em minus 0.4em\relax IEEE, 2011, pp.
  790--795.

\bibitem{hofmarcher2019visual}
M.~Hofmarcher, T.~Unterthiner, J.~Arjona-Medina, G.~Klambauer, S.~Hochreiter,
  and B.~Nessler, ``Visual scene understanding for autonomous driving using
  semantic segmentation,'' in \emph{Explainable AI: Interpreting, Explaining
  and Visualizing Deep Learning}.\hskip 1em plus 0.5em minus 0.4em\relax
  Springer, 2019, pp. 285--296.

\bibitem{xiang2018posecnn}
Y.~Xiang, T.~Schmidt, V.~Narayanan, and D.~Fox, ``{PoseCNN}: A convolutional
  neural network for {6D} object pose estimation in cluttered scenes,'' in
  \emph{Robotics: Science and Systems (RSS)}, 2018.

\bibitem{bloesch2015robust}
M.~Bloesch, S.~Omari, M.~Hutter, and R.~Siegwart, ``Robust visual inertial
  odometry using a direct ekf-based approach,'' in \emph{2015 IEEE/RSJ
  international conference on intelligent robots and systems (IROS)}.\hskip 1em
  plus 0.5em minus 0.4em\relax IEEE, 2015, pp. 298--304.

\bibitem{sola2022wolf}
J.~Sol{\`a}, J.~Vallv{\'e}, J.~Casals, J.~Deray, M.~Fourmy, D.~Atchuthan,
  A.~Corominas-Murtra, and J.~Andrade-Cetto, ``Wolf: A modular estimation
  framework for robotics based on factor graphs,'' \emph{IEEE Robotics and
  Automation Letters}, vol.~7, no.~2, pp. 4710--4717, 2022.

\bibitem{brommer2020mars}
C.~Brommer, R.~Jung, J.~Steinbrener, and S.~Weiss, ``{MaRS: A modular and
  robust sensor-fusion framework},'' \emph{IEEE Robotics and Automation
  Letters}, vol.~6, no.~2, pp. 359--366, 2020.

\bibitem{moore2016generalized}
T.~Moore and D.~Stouch, ``A generalized extended kalman filter implementation
  for the robot operating system,'' in \emph{Intelligent autonomous systems
  13}.\hskip 1em plus 0.5em minus 0.4em\relax Springer, 2016, pp. 335--348.

\bibitem{cao2016real}
Z.~Cao, Y.~Sheikh, and N.~K. Banerjee, ``Real-time scalable 6dof pose
  estimation for textureless objects,'' in \emph{2016 IEEE International
  conference on Robotics and Automation (ICRA)}.\hskip 1em plus 0.5em minus
  0.4em\relax IEEE, 2016, pp. 2441--2448.

\bibitem{pavlakos20176}
G.~Pavlakos, X.~Zhou, A.~Chan, K.~G. Derpanis, and K.~Daniilidis, ``6-dof
  object pose from semantic keypoints,'' in \emph{2017 IEEE international
  conference on robotics and automation (ICRA)}.\hskip 1em plus 0.5em minus
  0.4em\relax IEEE, 2017, pp. 2011--2018.

\bibitem{amini2021t6d}
A.~Amini, A.~S. Periyasamy, and S.~Behnke, ``T6d-direct: Transformers for
  multi-object 6d pose direct regression,'' in \emph{DAGM German Conference on
  Pattern Recognition}.\hskip 1em plus 0.5em minus 0.4em\relax Springer, 2021,
  pp. 530--544.

\bibitem{billings2019silhonet}
G.~Billings and M.~Johnson-Roberson, ``Silhonet: An rgb method for 6d object
  pose estimation,'' \emph{IEEE Robotics and Automation Letters}, vol.~4,
  no.~4, pp. 3727--3734, 2019.

\bibitem{kehl2017ssd}
W.~Kehl, F.~Manhardt, F.~Tombari, S.~Ilic, and N.~Navab, ``Ssd-6d: Making
  rgb-based 3d detection and 6d pose estimation great again,'' in
  \emph{Proceedings of the IEEE international conference on computer vision},
  2017, pp. 1521--1529.

\bibitem{li2018deepim}
Y.~Li, G.~Wang, X.~Ji, Y.~Xiang, and D.~Fox, ``Deepim: Deep iterative matching
  for 6d pose estimation,'' in \emph{Proceedings of the European Conference on
  Computer Vision (ECCV)}, 2018, pp. 683--698.

\bibitem{li2019cdpn}
Z.~Li, G.~Wang, and X.~Ji, ``Cdpn: Coordinates-based disentangled pose network
  for real-time rgb-based 6-dof object pose estimation,'' in \emph{Proceedings
  of the IEEE/CVF International Conference on Computer Vision}, 2019, pp.
  7678--7687.

\bibitem{wang2021gdr}
G.~Wang, F.~Manhardt, F.~Tombari, and X.~Ji, ``Gdr-net: Geometry-guided direct
  regression network for monocular 6d object pose estimation,'' in
  \emph{Proceedings of the IEEE/CVF Conference on Computer Vision and Pattern
  Recognition}, 2021, pp. 16\,611--16\,621.

\bibitem{li2018unified}
C.~Li, J.~Bai, and G.~D. Hager, ``A unified framework for multi-view
  multi-class object pose estimation,'' in \emph{Proceedings of the european
  conference on computer vision (eccv)}, 2018, pp. 254--269.

\bibitem{labbe2020cosypose}
Y.~Labb{\'e}, J.~Carpentier, M.~Aubry, and J.~Sivic, ``Cosypose: Consistent
  multi-view multi-object 6d pose estimation,'' in \emph{European Conference on
  Computer Vision}.\hskip 1em plus 0.5em minus 0.4em\relax Springer, 2020, pp.
  574--591.

\bibitem{krull2015learning}
A.~Krull, E.~Brachmann, F.~Michel, M.~Y. Yang, S.~Gumhold, and C.~Rother,
  ``Learning analysis-by-synthesis for 6d pose estimation in rgb-d images,'' in
  \emph{Proceedings of the IEEE international conference on computer vision},
  2015, pp. 954--962.

\bibitem{michel2017global}
F.~Michel, A.~Kirillov, E.~Brachmann, A.~Krull, S.~Gumhold, B.~Savchynskyy, and
  C.~Rother, ``Global hypothesis generation for 6d object pose estimation,'' in
  \emph{Proceedings of the IEEE Conference on Computer Vision and Pattern
  Recognition}, 2017, pp. 462--471.

\bibitem{jantos2023poet}
T.~Jantos, M.~A. Hamdad, W.~Granig, S.~Weiss, and J.~Steinbrener, ``{PoET: Pose
  Estimation Transformer for Single-View, Multi-Object 6D Pose Estimation},''
  in \emph{Proceedings of the 6th Conference on Robot Learning}.\hskip 1em plus
  0.5em minus 0.4em\relax PMLR, 2023.

\bibitem{salas2013slam++}
R.~F. Salas-Moreno, R.~A. Newcombe, H.~Strasdat, P.~H. Kelly, and A.~J.
  Davison, ``Slam++: Simultaneous localisation and mapping at the level of
  objects,'' in \emph{Proceedings of the IEEE conference on computer vision and
  pattern recognition}, 2013, pp. 1352--1359.

\bibitem{merrill2022symmetry}
N.~Merrill, Y.~Guo, X.~Zuo, X.~Huang, S.~Leutenegger, X.~Peng, L.~Ren, and
  G.~Huang, ``Symmetry and uncertainty-aware object slam for 6dof object pose
  estimation,'' in \emph{Proceedings of the IEEE/CVF Conference on Computer
  Vision and Pattern Recognition}, 2022, pp. 14\,901--14\,910.

\bibitem{bowman2017probabilistic}
S.~L. Bowman, N.~Atanasov, K.~Daniilidis, and G.~J. Pappas, ``Probabilistic
  data association for semantic slam,'' in \emph{2017 IEEE international
  conference on robotics and automation (ICRA)}.\hskip 1em plus 0.5em minus
  0.4em\relax IEEE, 2017, pp. 1722--1729.

\bibitem{thomas2015visual}
J.~Thomas, G.~Loianno, K.~Daniilidis, and V.~Kumar, ``Visual servoing of
  quadrotors for perching by hanging from cylindrical objects,'' \emph{IEEE
  robotics and automation letters}, vol.~1, no.~1, pp. 57--64, 2015.

\bibitem{loianno2018localization}
G.~Loianno, V.~Spurny, J.~Thomas, T.~Baca, D.~Thakur, D.~Hert, R.~Penicka,
  T.~Krajnik, A.~Zhou, A.~Cho, \emph{et~al.}, ``Localization, grasping, and
  transportation of magnetic objects by a team of mavs in challenging
  desert-like environments,'' \emph{IEEE Robotics and Automation Letters},
  vol.~3, no.~3, pp. 1576--1583, 2018.

\bibitem{mathe2016vision}
K.~M{\'a}th{\'e}, L.~Bu{\c{s}}oniu, L.~Barab{\'a}s, C.-I. Iuga, L.~Miclea, and
  J.~Braband, ``Vision-based control of a quadrotor for an object inspection
  scenario,'' in \emph{2016 International Conference on Unmanned Aircraft
  Systems (ICUAS)}.\hskip 1em plus 0.5em minus 0.4em\relax IEEE, 2016, pp.
  849--857.

\bibitem{sola2017quaternion}
J.~Sola, ``Quaternion kinematics for the error-state kalman filter,''
  \emph{arXiv preprint arXiv:1711.02508}, 2017.

\bibitem{brommer2020improved}
C.~Brommer, C.~B{\"o}hm, J.~Steinbrener, R.~Brockers, and S.~Weiss, ``Improved
  state estimation in distorted magnetic fields,'' in \emph{2020 International
  Conference on Unmanned Aircraft Systems (ICUAS)}.\hskip 1em plus 0.5em minus
  0.4em\relax IEEE, 2020, pp. 1007--1013.

\bibitem{furgale2013unified}
P.~Furgale, J.~Rehder, and R.~Siegwart, ``Unified temporal and spatial
  calibration for multi-sensor systems,'' in \emph{2013 IEEE/RSJ International
  Conference on Intelligent Robots and Systems}.\hskip 1em plus 0.5em minus
  0.4em\relax IEEE, 2013, pp. 1280--1286.

\end{thebibliography}
\end{document}